\documentclass[acmtog]{acmart}

\usepackage{booktabs} 

\colorlet{blue}{black}

\usepackage{multirow}   
\usepackage{booktabs}

\usepackage[ruled]{algorithm2e} 

\SetAlFnt{\small}
\SetAlCapFnt{\small}
\SetAlCapNameFnt{\small}
\SetAlCapHSkip{0pt}

\acmJournal{TOG}

\copyrightyear{2026}
\acmYear{2026}
\setcopyright{cc}
\setcctype{by}
\acmConference[SA Conference Papers '26]{SIGGRAPH Asia 2026 Conference Papers}{December 01--04, 2026}{Kuala Lumpur, Malaysia}
\acmBooktitle{SIGGRAPH Asia 2026 Conference Papers (SA Conference Papers '26), December 01--04, 2026, Kuala Lumpur, Malaysia}
\acmDOI{10.1145/3829340.3842167}
\acmISBN{979-8-4007-2842-6/2026/12}

\begin{document}
\title{CGGT: Curve-Grounded Geometry Transformer for 3D Parametric Curve Reconstruction}


\author{Zhirui Gao}
\orcid{0000-0002-7108-7962}
\authornote{Co-first author}
\affiliation{%
  \institution{National University of Defense Technology}
  \city{Changsha}
  \country{China}}
\email{gzrer2018@gmail.com}

\author{Renjiao Yi}
\orcid{0000-0002-6057-1089}
\authornotemark[1]
\affiliation{%
  \institution{National University of Defense Technology}
  \city{Changsha}
  \country{China}}
\email{yirenjiao@nudt.edu.cn}

\author{Yunfan Ye}
\orcid{0000-0001-5723-9627}
\affiliation{%
  \institution{Hunan University}
  \city{Changsha}
  \country{China}}
\email{yeyunfan@hnu.edu.cn}

\author{Ruizhen Hu}
\orcid{0000-0002-6798-0336}
\affiliation{%
  \institution{Shenzhen University}
  \city{Shenzhen}
  \country{China}}
\email{ruizhen.hu@gmail.com}

\author{Chenyang Zhu}
\orcid{0000-0003-2838-8601}
\affiliation{%
  \institution{National University of Defense Technology}
  \city{Changsha}
  \country{China}}
\email{zhuchenyang07@nudt.edu.cn}

\author{Wei Chen}
\orcid{0000-0002-2876-6687}
\authornote{Corresponding author  \\ Project page: \url{https://zhirui-gao.github.io/cggt/}}
\affiliation{%
  \institution{National University of Defense Technology}
  \city{Changsha}
  \country{China}}
\email{chenwei@nudt.edu.cn}

\author{Kai Xu}
\orcid{0000-0002-9054-0216}
\authornotemark[2]

\affiliation{%
  \department{Jiangsu Key Laboratory of AI for Industries}
  \institution{Institute of AI for Industries, Chinese Academy of Sciences}
  \city{Nanjing}
  \country{China}}
\email{kevin.kai.xu@gmail.com}

\renewcommand{\shortauthors}{Gao et al.}

%
%


\begin{teaserfigure}
  \includegraphics[width=0.97\textwidth]{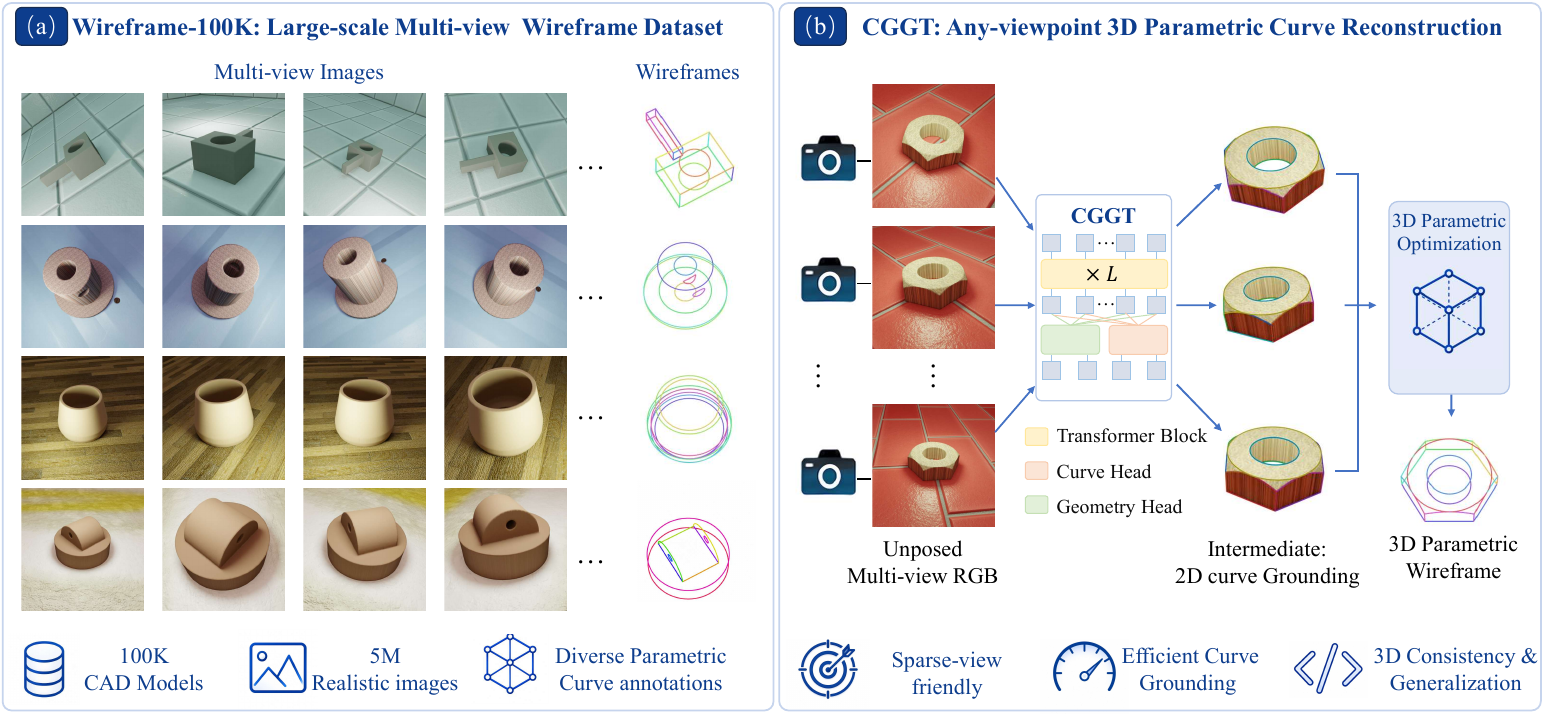}
  \vspace{-8pt}
  \caption{
  (a) We introduce {Wireframe-100K}, a curated large-scale dataset containing 100K CAD models, 5M realistic images, and diverse parametric curve annotations. 
  (b)  We propose CGGT, a geometry-aware transformer for grounding 3D-consistent 2D curve instances from sparse, unposed multi-view images, followed by fast 3D parametric optimization for editable CAD wireframe reconstruction.}
  \Description{} 
  \label{fig:teaser}
\end{teaserfigure}
\begin{abstract}

Recovering editable 3D parametric curves from 2D images is a fundamental challenge in computer graphics, 
bridging pixel-based perception and vector-based CAD modeling. Existing NeRF- and 3DGS-based methods often rely on dense calibrated views, precomputed 2D edge maps, and costly per-scene optimization, limiting their applicability to casually captured real-world inputs. We propose \textit{\textbf{CGGT}}, a \textit{\textbf{C}urve-\textbf{G}rounded \textbf{G}eometry \textbf{T}ransformer} that directly grounds 3D-consistent 2D curve instances in the image space from sparse, unposed multi-view images. CGGT combines a geometry-aware transformer encoder for multi-view feature learning with a curve-aware masked-attention decoder for cross-view instance association. In a single forward pass, it predicts camera parameters, dense depth maps, and instance-level 2D curve masks, which are then lifted into 3D and refined through a fast parametric optimization stage to recover compact, editable 3D curve primitives. To support structured curve learning, we introduce Wireframe-100K, a large-scale dataset comprising 100,000 CAD models with diverse topologies, realistic multi-view renderings, and accurate parametric curve annotations. Extensive experiments show that our framework achieves substantial improvements in both reconstruction accuracy and efficiency, particularly under challenging sparse-view settings and in separating persistent 3D structural edges from view-dependent image edges caused by silhouettes, textures, and appearance variations. Despite being trained solely on synthetic data, CGGT generalizes well to real-world images, demonstrating its potential for practical CAD-style wireframe reconstruction from unconstrained visual inputs.

\end{abstract}

\maketitle

\section{Introduction}

Recovering editable 3D parametric curves from 2D images is a fundamental problem in computer vision and graphics. Parametric curves provide a compact, interpretable, and editable representation of object geometry, making them \textcolor{blue}{valuable} for CAD modeling, reverse engineering~\cite{choi2023app_cad_1, zhang2023app_cad_2, li2024app_cad_3}, industrial design, and augmented reality~\cite{qin2021lightautonomousdriving, cheng2021roadmapping}. Unlike dense point clouds, meshes, or neural fields, 3D curves explicitly encode structural elements such as lines, arcs, ellipses, and free-form boundaries~\cite{choi20243doodle}. \textcolor{blue}{Recent work explores structured representations for programmatic CAD construction~\cite{willis2021fusion}, primitive reconstruction and B-rep reconstruction/generation~\cite{li2023surface,liu2024split,xu2024brepgen,lee2025brepdiff}, and fabrication-aware reverse engineering~\cite{noeckel2021fabrication}. Accordingly, we position 3D parametric curves as an editable structural scaffold and an intermediate representation for downstream CAD/B-rep reconstruction.}

Despite their importance, reconstructing clean and topologically coherent 3D curves from multi-view images remains highly ill-posed. Recent methods~\cite{Ye_2023_CVPR, Li2024CVPR, chelani2024edgegaussians, yang2025sgcr} have explored neural rendering representations, such as NeRF~\cite{mildenhall2020nerf} and 3D Gaussian Splatting~\cite{kerbl20233dgs}, for curve reconstruction. These methods typically optimize scene-specific volumetric or point-based representations and then extract curve structures from the optimized geometry or learned edge-aware fields. Such reconstruction-then-extraction pipelines are prone to error accumulation, as inaccuracies in the intermediate scene representation can directly affect the extracted curves. More recent single-stage differentiable rendering methods, such as CurveGaussian~\cite{Gao_2025_ICCV} and SketchSplat~\cite{Ying_2025_ICCV}, provide a more elegant formulation by optimizing curve-aware primitives directly within the rendering process, improving efficiency and reducing intermediate extraction errors. However, they still rely on per-scene optimization. As a result, these methods often require dense calibrated views, reliable object masks or edge maps \cite{su2021pdc,ye2023crispness,ye2024diffusionedge}, and costly test-time fitting, limiting their applicability to casual real-world captures with sparse and unconstrained viewpoints.

A further challenge is that image-space edges are not equivalent to object-level geometric curves. Texture boundaries, shadows, silhouettes, and view-dependent occlusion contours can produce strong but misleading pseudo-edges. Conversely, true geometric curves may appear weak, fragmented, or partially occluded in individual views. Without object-level and multi-view priors, optimization-centric methods can easily overfit to such local edge evidence, leading to fragmented, redundant, or topologically inconsistent curves.

To address these limitations, we propose a two-stage reconstruction framework centered on the Curve-Grounded Geometry Transformer (CGGT). CGGT is a feed-forward curve grounding module that grounds 3D-consistent 2D curve instances in the image space from sparse, unposed multi-view images. Rather than learning 3D curve representations directly, CGGT leverages multi-view geometry priors from modern 3D foundation models, allowing the network to focus on curve grounding and cross-view instance association. Specifically, CGGT couples a geometry-aware transformer encoder~\cite{wang2025vggt} for multi-view feature extraction with a curve-aware masked-attention decoder for instance-level curve grounding. In a single forward pass, it jointly predicts camera parameters, dense depth maps, and multi-view-consistent 2D curve masks. The grounded curve pixels are subsequently lifted into 3D using the predicted camera parameters and depth maps, and fitted with parametric curve primitives. Finally, a fast test-time optimization refines these primitives by enforcing multi-view consistency and geometric regularity, producing clean, compact, and editable 3D parametric curves.

Training such a model requires large-scale supervision that couples realistic image observations with accurate curve-level geometry. To this end, we construct Wireframe-100K, a large-scale synthetic dataset containing 100,000 diverse CAD models~\cite{koch2019abc}, realistic multi-view renderings, and annotations of diverse parametric curve types. This dataset enables CGGT to learn object-level curve priors beyond local image edges, helping the model suppress pseudo-edges such as texture and silhouette edges and remain robust to large viewpoint changes and cluttered backgrounds.

Our main contributions are summarized as follows:
\begin{itemize}
\item We propose CGGT, a geometry-aware transformer module that directly grounds instance-level, multi-view consistent 2D curve masks from sparse, unposed images. 

\item We introduce Wireframe-100K, a large-scale dataset for data-driven 3D curve reconstruction, containing 100,000 diverse CAD models, realistic multi-view renderings, and accurate annotations of diverse parametric curve types.

\item We show that our framework supports sparse, unposed inputs without object-mask extraction or edge-map preprocessing, while mitigating pseudo-edges such as texture and silhouette edges. It achieves strong improvements in geometric accuracy, curve compactness, and reconstruction efficiency, with zero-shot generalization to real-world captures.
\end{itemize}

\section{Related Work}

\subsection{3D Parametric Curve Reconstruction}

Traditional SfM, silhouette-based reconstruction, and line-matching methods~\cite{schonberger2016structure,sinha2005multi,micusik2017structure,schindler2006line} rely on reliable cross-view correspondences and their performance is sensitive to sparse views, weak texture, and occlusions. \textcolor{blue}{To avoid directly lifting 2D curves into 3D, some methods detect sharp features from point clouds~\cite{himeur2021pcednet,yu2018ec,10.1145/3528223.3530140}. In contrast, others recover parametric edge primitives~\cite{wang2020pie,zhu2023nerve} or structured wireframes and CAD surfaces~\cite{liu2021pc2wf,feng2025d,Dong2024NeurCADRecon}. However, their performance depends strongly on the quality of the intermediate geometry.} \textcolor{blue}{Complementary studies have explored projected-line analysis, image primitive decomposition, and sketch-based modeling~\cite{yu2018projectedlines,huang2018deepprimitive,miao2015symmsketch}, as well as curve completion and structural curve modeling~\cite{ma2018centercurve}. Other methods construct B-spline curves with controlled curvature and knot placement~\cite{zhang2020computing}. These works provide useful geometric priors but do not address multi-view curve reconstruction from images.}

Recent neural rendering methods recover curves from multi-view images using NeRF~\cite{mildenhall2020nerf} or 3D Gaussian Splatting~\cite{kerbl20233dgs,huang20242dgs}. {NEF, EMAP, and NEAT reconstruct implicit edge or attraction fields before extracting explicit curve structures~\cite{Ye_2023_CVPR,Li2024CVPR,xue2024neat}. EdgeGaussians and SGCR encode edge structures using Gaussian primitives~\cite{chelani2024edgegaussians,yang2025sgcr}, while BrepGaussian and SeCuRe further target CAD-oriented or sparse-view reconstruction~\cite{yu2026brepgaussian,feng2026secure}.} More recent single-stage differentiable rendering methods, such as SketchSplat~\cite{Ying_2025_ICCV} and CurveGaussian~\cite{Gao_2025_ICCV}, directly optimize curve-aware primitives against image observations. Despite these advances, such methods remain optimization-centric and typically require dense calibrated views, reliable masks or edge maps, and costly per-scene fitting.


\subsection{ CAD Reconstruction and Generation}

\textcolor{blue}{ CAD methods parse geometric primitives from sketches~\cite{wang2024parametric}, recover B-rep geometry from point clouds~\cite{li2023surface,liu2024split,11493603}, or model CAD construction sequences and cross-model correspondences~\cite{willis2021fusion,jones2023brepmatching}. Learning-based approaches further model structured wireframes~\cite{xu2026wg,ma2025clr} and complete B-reps~\cite{GuoComplexGen2022,liu2025hola,li2025brepgpt}, including recent diffusion-based models~\cite{xu2024brepgen,lee2025brepdiff,sun2025extreme,10901942}. Fabrication-aware methods incorporate manufacturing constraints into geometric reconstruction and design~\cite{noeckel2021fabrication,wu2024tune}, while MIND performs inverse design conditioned on target physical properties~\cite{10.1145/3721238.3730682}. These directions are summarized in a recent survey of AI-driven CAD generation~\cite{wu2026ai}.}

\begin{figure*}[t]
  \centering
  \includegraphics[width=0.98\textwidth]{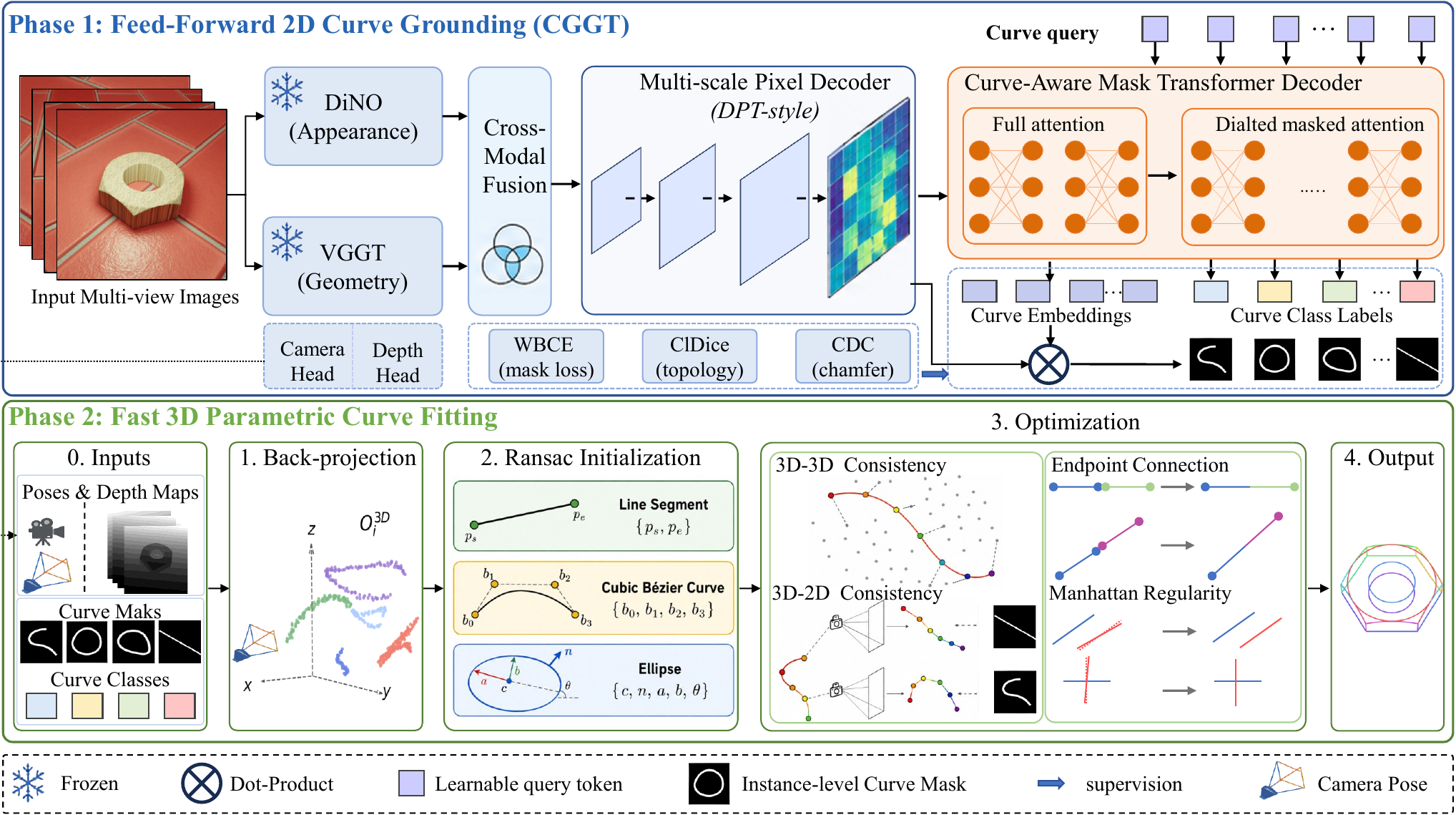}
  \vspace{-5pt}
  \caption{\textbf{Method overview}. Our framework consists of two stages: 
  (i) a feed-forward curve grounding module that identifies and localizes curve-instance pixels in the image space, and 
  (ii) \textcolor{blue}{a fast parametric optimization stage that lifts the grounded curve pixels into 3D and fits them with compact parametric curve primitives.}}
  \label{fig:method-pipeline}
\end{figure*}

\subsection{Spatial Foundation Models}

Image-to-3D reconstruction has long been a central problem in computer vision. Classical pipelines such as COLMAP~\cite{schonberger2016structure} recover scene structure through sparse feature matching and geometric verification. Neural representations, including NeRF~\cite{mildenhall2020nerf} and 3DGS~\cite{kerbl20233dgs}, achieve high-quality novel view synthesis but typically require accurate camera poses and costly per-scene optimization. More recently, spatial foundation models~\cite{wang2024dust3r,wang2025vggt,depthanything3} have enabled feed-forward 3D perception from unposed images by regressing dense point maps, depth, or latent geometry. These models have further been extended to spatial perception~\cite{fan2025vlm, qu2025spatialvla, chenfeedforward, 11366109, 11153745} and semantic scene understanding~\cite{iggt2024,10901941}. In this work, we adapt such geometry-aware representations to a more structured setting: grounding multi-view consistent curve instances and supporting 3D parametric curve reconstruction.
\section{Method}

Given unposed and unordered RGB images
$\mathcal{I}=\{I_1,\dots,I_N\}$, our goal is to reconstruct editable 3D parametric curves
$\mathcal{C}=\{C_1,\dots,C_K\}$ representing the CAD wireframe of the object.
The key challenge is to identify curve instances in each view, associate them across views, and recover their continuous 3D geometry.

We address this problem with a two-stage framework.
The first stage, CGGT, grounds 3D-consistent 2D curve instances in image space and predicts the camera and depth information needed to lift them into 3D.
The second stage fits parametric curve primitives from the lifted curve points through fast test-time optimization, enforcing multi-view consistency and geometric regularity. At a high level, the desired reconstruction can be expressed as:
\begin{equation}
    \min_{\Theta, \mathcal{C}}
    \underbrace{
    \mathcal{D}_{\mathrm{mask}}
    \left(
    \hat{\mathcal{M}},
    \Pi(\mathcal{C}^{*})
    \right)
    }_{\text{2D curve grounding}}
    +
    \lambda
    \underbrace{
    \mathcal{D}_{\mathrm{curve}}
    \left(
    \hat{\mathcal{C}},
    \mathcal{C}^{*}
    \right)
    }_{\text{3D curve reconstruction}},
\end{equation}
where $\Theta$ denotes the parameters of CGGT, 
$\hat{\mathcal{M}}$ is the predicted multi-view 2D curve masks, 
$\Pi(\mathcal{C}^{*})$ is the projection of the ground-truth 3D curves onto the input views, 
$\hat{\mathcal{C}}$ is the final reconstructed 3D parametric curves, and 
$\mathcal{C}^{*}$ is the ground-truth curve set.

\subsection{Architecture of CGGT}

CGGT is the feed-forward curve grounding module in the first stage of our framework. 
Given $N$ unposed input images, CGGT predicts camera parameters, dense depth maps, instance-level curve masks, and curve categories in a single forward pass. 
As shown in Fig.~\ref{fig:method-pipeline}, CGGT consists of three components: 
(1) a \textbf{Large Unified Transformer Encoder}, which extracts and fuses pretrained geometry-aware and appearance-aware tokens; 
(2) a \textbf{Multi-scale Pixel Decoder}, which converts the fused tokens into dense multi-resolution feature maps with depth-aware refinement; and 
(3) a \textbf{Curve-Aware Mask Transformer Decoder}, which formulates curve grounding as query-based set prediction and outputs cross-view consistent curve masks and categories.

\subsubsection{Large Unified Transformer Encoder}

To obtain robust multi-view representations, we build our encoder upon VGGT~\cite{wang2025vggt}, a 3D foundation model for feed-forward reconstruction from unposed multi-view images. Given $N$ images, VGGT encodes each view into a sequence of geometry-aware tokens
$\{T_{\mathrm{vggt}}^{i} \in \mathbb{R}^{M \times D_g}\}_{i=1}^{N}$, where $M$ denotes the number of tokens per image and $D_g$ is the feature dimension.  While VGGT provides strong priors for camera estimation, depth prediction, and 3D structure reasoning, its tokens are primarily optimized for global reconstruction and may lose fine image details needed for curve localization. 
To complement them with local appearance cues, we extract appearance-aware tokens from a frozen DINOv2 encoder~\cite{oquab2023dinov2}, denoted as
$\{T_{\mathrm{dino}}^{i} \in \mathbb{R}^{M \times D_a}\}_{i=1}^{N}$.
We then fuse the geometry-aware VGGT tokens and appearance-aware DINO tokens using a lightweight cross-attention module:
\begin{equation}
\tilde{T}_{i}
=
\mathrm{CrossAttn}
\left(
Q = T_{\mathrm{vggt}}^{i},
K = T_{\mathrm{dino}}^{i},
V = T_{\mathrm{dino}}^{i}
\right).
\end{equation}
In this design, VGGT tokens query DINO tokens to selectively retrieve appearance details while preserving multi-view geometric consistency. 
The fused tokens $\tilde{T}_{i}$ are therefore both geometry-aware and appearance-sensitive, which is important for grounding thin and weakly visible curve structures in image space. We apply this fusion to the intermediate feature layers used by the subsequent DPT-style decoder, producing multi-level curve tokens
$\boldsymbol{T}^{\mathrm{curve}}_i=\{\tilde{T}_{i}^{(l)}\}_{l=1}^{4}$.

\subsubsection{Multi-scale Pixel Decoder}

The pixel decoder converts the fused multi-level tokens 
$\boldsymbol{T}^{\mathrm{curve}}_i=\{\tilde{T}_{i}^{(l)}\}_{l=1}^{4}$ 
into dense feature maps for fine-grained curve mask prediction. 
We adopt a DPT-style decoder~\cite{ranftl2021vision}: each token feature $\tilde{T}_{i}^{(l)}$ is projected to a common channel dimension, reshaped according to the encoder patch layout, and progressively fused in a top-down manner. 
This produces a hierarchy of dense curve-aware features:
\begin{equation}
\boldsymbol{F}^{\mathrm{curve}}_i
=
\{F^{\mathrm{curve}}_{i,(l)}\}_{l=1}^{4}.
\end{equation}

\paragraph{Depth-Aware Enhancement.}
Depth cues are particularly useful for localizing curves near depth discontinuities, occlusion boundaries, and sharp geometric changes. 
We therefore enhance the dense curve features with depth features from the frozen VGGT depth head. 
At each scale, we combine the visual curve feature and the corresponding depth feature using a lightweight MLP-based fusion:
\begin{equation}
    \hat{F}^{\mathrm{curve}}_{i,(l)}
    =
    \mathrm{MLP}_{\mathrm{curve}}
    \left(F^{\mathrm{curve}}_{i,(l)}\right)
    +
    \mathrm{MLP}_{\mathrm{depth}}
    \left(F^{\mathrm{depth}}_{i,(l)}\right).
\end{equation}
The enhanced features encode both image-level curve evidence and depth-aware geometric structure. The pixel decoder outputs a multi-scale feature pyramid and a high-resolution per-pixel embedding:
\begin{equation}
\begin{split}
\mathcal{F} = \{&
F_1 \in \mathbb{R}^{C_1 \times 2h \times 2w}, \;
F_2 \in \mathbb{R}^{C_2 \times 4h \times 4w}, \\
&
F_3 \in \mathbb{R}^{C_3 \times 8h \times 8w}, \;
\epsilon_{\mathrm{pixel}} \in \mathbb{R}^{C_\epsilon \times H_\epsilon \times W_\epsilon}
\},
\end{split}
\end{equation}
where $h$ and $w$ are the spatial resolution of the encoder token grid. 
The multi-scale features $\{F_1,F_2,F_3\}$ serve as keys and values in the curve decoder, while $\epsilon_{\mathrm{pixel}}$ is used for final mask prediction.

\subsubsection{Curve-Aware Mask Transformer Decoder}

We formulate curve grounding as a query-based set prediction task. 
The decoder takes the multi-scale feature pyramid $\{F_1,F_2,F_3\}$ and a fixed set of learnable curve queries, and predicts curve embeddings $E_{\mathrm{curve}}$ for instance-level mask and category prediction. 
Following Mask2Former~\cite{cheng2022masked}, the decoder is composed of $L$ repeated layers that attend to multi-scale image features. 
We aggregate features from all views into a view-indexed feature sequence, enabling each curve query to collect cross-view evidence for the same curve instance.

A standard Mask2Former decoder uses masked cross-attention, where each query attends only to the foreground region predicted by the previous layer. 
However, this strategy is fragile for curve grounding, since curves occupy only a small fraction of image pixels and early mask predictions are often incomplete or noisy. 
Using such masks directly to restrict attention can prematurely suppress thin curve structures, making them difficult to recover in later layers.

We therefore introduce a curve-aware attention strategy. 
In the first two decoder layers, we use full cross-attention, allowing curve queries to explore the entire feature map before reliable masks are available. 
For subsequent layers, we apply masked cross-attention with dilated masks from the previous layer. 
The dilation expands the attended regions around predicted curves, reducing the risk of missing thin structures while suppressing irrelevant background features. For each curve query, the final mask is obtained by taking the dot product between its mask embedding and the dense per-pixel embedding. In addition, a lightweight classification head predicts the corresponding curve category, including Line, Bézier, Ellipse, and Null/No-Curve.  


\subsection{Training  of CGGT}

To learn sparse curve structures while preserving geometric consistency, we train CGGT with a composite objective and a curve-biased point sampling strategy.

\subsubsection{Loss Functions}

\paragraph{Class-Balanced Weighted BCE.}
Curve pixels occupy only a small fraction of the image, making standard binary cross-entropy biased toward the background. 
We therefore adopt a weighted binary cross-entropy (WBCE) loss to balance foreground and background contributions. 
Let $p_i$ be the predicted probability and $y_i \in \{0,1\}$ be the ground-truth label at pixel $i$. 
The loss is defined as
\begin{equation}
    \mathcal{L}_{\mathrm{wbce}} =
    - \sum_i
    \left[
    \beta y_i \log(p_i)
    +
    (1-\beta)(1-y_i)\log(1-p_i)
    \right],
\end{equation}
where $\beta = |Y_-| / (|Y_+| + |Y_-|)$ increases the relative weight of sparse curve pixels $Y_+$ against the dominant background pixels $Y_-$.

\paragraph{Topological Consistency.}
Thin curves are prone to fragmentation. To enforce connectivity, we incorporate topology-preserving Dice loss (clDice~\cite{shit2021cldice}). Unlike area-based metrics~\cite{milletari2016v}, clDice penalizes topological errors by measuring the overlap between the morphological skeletons of the predicted and ground-truth masks, ensuring geometric continuity.

\paragraph{Curve-Depth Consistency}
3D geometric curves typically reside on sharp geometric boundaries. While depth values may change abruptly across an occlusion edge, the depth along the curve itself should be locally smooth. To utilize this geometric prior effectively, we introduce a Curve-Depth Consistency (CDC) loss using the frozen depth map $D$. We penalize adjacent pixels that are both predicted as belonging to the same curve instance but exhibit large depth discrepancies. Such discrepancies typically indicate false positive predictions bridging foreground and background, which would otherwise introduce significant noise and outliers to the subsequent 3D parametric curve fitting process. The loss is defined as:
\begin{equation}
    \mathcal{L}_{\text{cdc}} = \sum_{k} \sum_{i} \sum_{j \in \mathcal{N}(i)} (p_{k,i} \cdot p_{k,j}) \cdot |D_i - D_j|,
\end{equation}
where $p_{k,i}$ and $p_{k,j}$ denote the predicted probabilities of pixel $i$ and its neighbor $j \in \mathcal{N}(i)$ belonging to the $k$-th curve instance, and $D_i, D_j$ are their corresponding depth values. This term acts as a regularizer to suppress noisy predictions that violate local 3D continuity.

 The total loss is defined as:
\begin{equation}
    \mathcal{L}_{\text{total}} = \lambda_{\text{wbce}}\mathcal{L}_{\text{wbce}} + \lambda_{\text{cldice}}\mathcal{L}_{\text{cldice}} + \lambda_{\text{cdc}}\mathcal{L}_{\text{cdc}}.
\end{equation}

\subsubsection{Improving Training Efficiency}

Following Mask2Former~\cite{cheng2022masked}, we compute the losses on a set of sampled points instead of the full image grid. 
However, uniform point sampling is inefficient for curve masks because curve pixels are extremely sparse. 
We therefore adopt a curve-biased sampling strategy, where points are sampled more densely around ground-truth curves and their boundaries.  This focuses supervision on informative regions, improves foreground-background balance, and accelerates convergence.

\section{Parametric Curve Optimization}

The grounded 2D curve masks are lifted into 3D using the predicted depth maps and camera poses. 
However, the lifted points may contain noise from depth errors, pose inaccuracies, and imperfect masks. 
We therefore perform a fast test-time optimization to refine parametric curve primitives. 
As illustrated in Fig.~\ref{fig:method-pipeline}, we first initialize each detected curve instance by RANSAC-based fitting on the lifted 3D points, and then optimize its parameters to enforce multi-view consistency and geometric regularity. 
For the $i$-th curve, we denote its parameters as $\mathcal{P}_i$: a line segment is represented by endpoints $\{\mathbf{p}_s,\mathbf{p}_e\}$, a cubic Bézier curve by control points $\{\mathbf{b}_0,\mathbf{b}_1,\mathbf{b}_2,\mathbf{b}_3\}$, and an ellipse by $\{\mathbf{c},\mathbf{n},a,b,\theta\}$.

\subsection{Geometric Consistency}

For each parametric curve $\mathcal{P}_i$, we uniformly sample 3D points $\mathcal{S}_i$ and align them with both lifted 3D observations and image-space curve evidence:
\begin{equation}
    \mathcal{L}_{\mathrm{cons}} =
    \sum_i
    \left[
    \mathrm{CD}_{3D}
    \left(
    \mathcal{S}_i,
    \mathcal{O}^{3D}_i
    \right)
    +
    \lambda_{2D}
    \sum_{v=1}^{V}
    \mathrm{CD}_{2D}
    \left(
    \pi_v(\mathcal{S}_i),
    \mathcal{M}^{v}_i
    \right)
    \right],
\end{equation}
where $\mathcal{O}^{3D}_i$ denotes the initially lifted 3D points, $\mathcal{M}^{v}_i$ denotes the grounded 2D curve pixels in view $v$, and $\pi_v(\cdot)$ is the projection function. 
The 3D term fits the lifted observations, while the 2D term enforces multi-view reprojection consistency.

\subsection{Geometric Regularization}

To improve connectivity and structural regularity, we introduce two lightweight priors.

\paragraph{Endpoint Connection.}
To close small gaps between adjacent curves, we apply a snapping regularizer~\cite{Li2024CVPR,Gao_2025_ICCV} over all endpoints $\mathcal{E}$:
\begin{equation}
    \mathcal{L}_{\mathrm{conn}} =
    \sum_{\mathbf{x}, \mathbf{y} \in \mathcal{E}, \mathbf{x} \neq \mathbf{y}}
    \mathbb{I}
    \left(
    \|\mathbf{x}-\mathbf{y}\|_2 < \tau
    \right)
    \|\mathbf{x}-\mathbf{y}\|_2^2 .
\end{equation}

\paragraph{Manhattan Regularity.}
For man-made objects, we encourage curve pairs that are already nearly parallel or orthogonal to satisfy these relations better. 
Let $\mathbf{v}_i$ and $\mathbf{v}_j$ be principal vectors, such as line directions or ellipse normals. 
We define
\begin{equation}
    \Omega_{\parallel}
    =
    \{(i,j)\mid |\mathbf{v}_i \cdot \mathbf{v}_j| > 1-\delta \},
    \quad
    \Omega_{\perp}
    =
    \{(i,j)\mid |\mathbf{v}_i \cdot \mathbf{v}_j| < \delta \}.
\end{equation}
The conditional Manhattan regularizer~\cite{coughlan2000manhattan} is
\begin{equation}
    \mathcal{L}_{\mathrm{man}} =
    \sum_{(i,j)\in\Omega_{\parallel}}
    \left(1-|\mathbf{v}_i \cdot \mathbf{v}_j|\right)
    +
    \sum_{(i,j)\in\Omega_{\perp}}
    |\mathbf{v}_i \cdot \mathbf{v}_j| .
\end{equation}

The final optimization objective is
\begin{equation}
    \mathcal{L}_{\mathrm{opt}}
    =
    \mathcal{L}_{\mathrm{cons}}
    +
    \lambda_{\mathrm{reg}}
    \left(
    \mathcal{L}_{\mathrm{conn}}
    +
    \mathcal{L}_{\mathrm{man}}
    \right).
\end{equation}

\begin{figure*}[t!]
    \centering
  \includegraphics[width=0.98\linewidth]{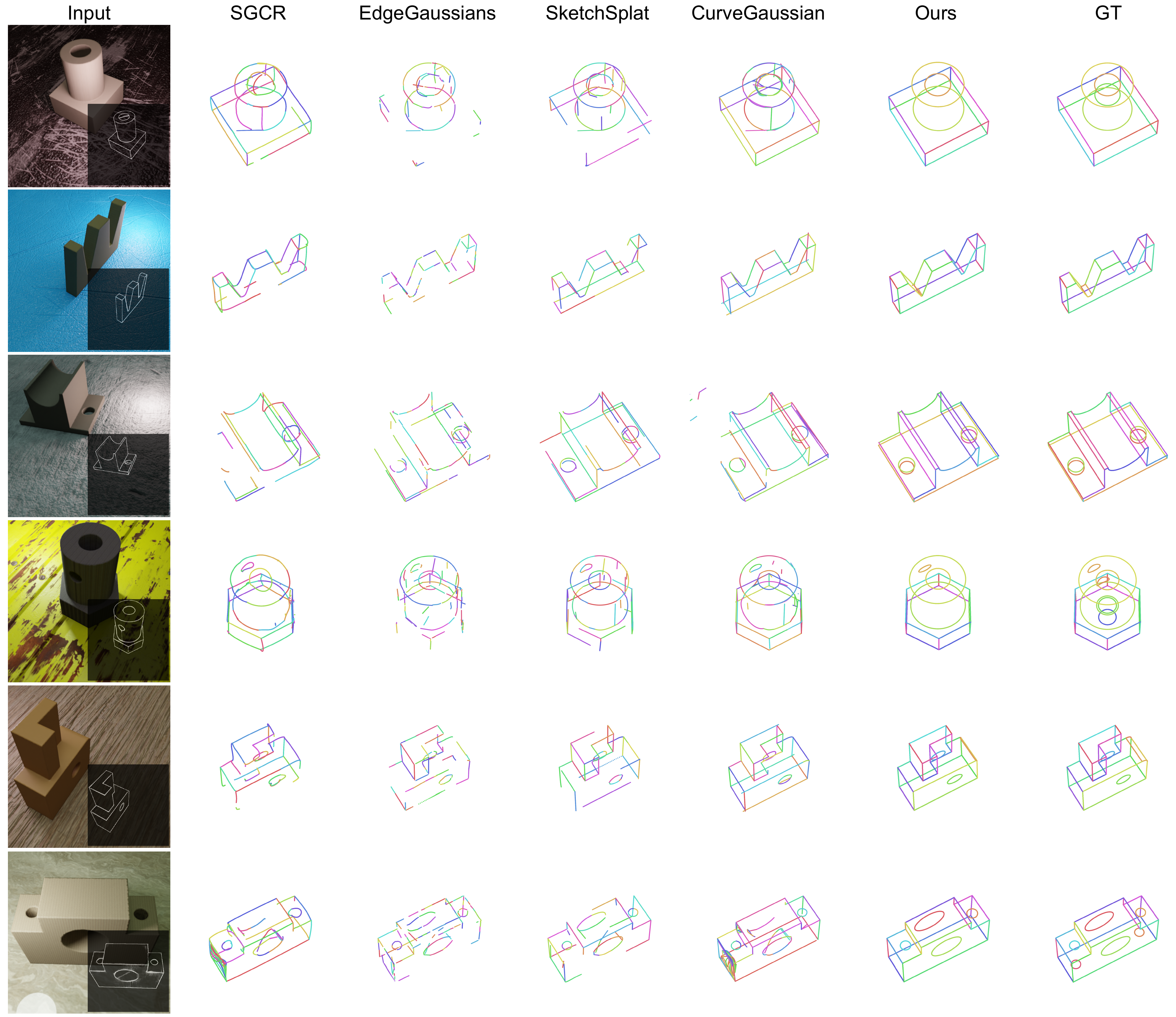}
  \vspace{-8pt}
   \caption{\textbf{Qualitative comparison on the ABC-NEF dataset~\cite{Ye_2023_CVPR}.}
Compared with state-of-the-art baselines, our method achieves higher reconstruction accuracy and better structural fidelity. Unlike baselines that rely on preprocessed edge images, our method directly takes the original images.}
  \label{fig:fig-main}
  \end{figure*}

\newcommand{\cmark}{\textcolor{green!70!black}{\checkmark}} 
\newcommand{\xmark}{\textcolor{red}{\times}}       

\begin{table*}[t]
\centering
\caption{\textbf{Quantitative results on ABC-NEF dataset~\cite{Ye_2023_CVPR}.} We compare our method against state-of-the-art approaches. The \textbf{Edge-free} column indicates whether a method relies on object edge detectors ($\xmark$) or requires no edge-detection preprocessing ($\cmark$). \# Image denotes the number of input images.}
\vspace{-5pt}

\label{tab:ABC}
\resizebox{0.98\linewidth}{!}{
\begin{tabular}{c|c|c|cc|ccc|ccc|ccc}
    \hline
    Method &
    Edge-free &
    \# Image &
    Acc.$\downarrow$ &
    Comp.$\downarrow$ &
    $\text{R}_{10} \uparrow$ &
    $\text{R}_{20}\uparrow$ &
    $\text{R}_{30} \uparrow$ &
    $\text{P}_{10} \uparrow$ &
    $\text{P}_{20} \uparrow$ &
    $\text{P}_{30} \uparrow$ &
    $\text{F}_{10} \uparrow$ &
    $\text{F}_{20} \uparrow$ &
    $\text{F}_{30}\uparrow$ \\ \hline

    \multirow{2}{*}{EMAP} &
    \multirow{2}{*}{$\xmark$} &
    15   &39.6
     &111.3 &43.6&56.4&61.7&62.6&77.5&81.7&49.7&59.5&66.0\\
    & &
    50   &23.5
     &45.9 &52.6&64.6&68.5&67.0&82.3&83.1&53.6&67.6&74.2\\ \hline

    \multirow{3}{*}{EdgeGaussians} &
    \multirow{3}{*}{$\xmark$} &
    5   &14.1
     &155.4 &3.2&9.6&16.3&82.8&90.7&87.8&27.5&33.6&38.3\\
    & &
    15   &11.2
     &71.3 &32.8&42.6&47.9&87.0&92.4&93.8&52.2&62.6&66.2\\
    & &
    50   &17.7
     &48.5&53.7&62.0&66.4&82.8&87.2&88.2&63.6&70.7&74.0\\ \hline

    \multirow{3}{*}{SGCR} &
    \multirow{3}{*}{$\xmark$} &
    5   &57.6
     &55.1 &38.8&50.7&57.1&40.7&56.3&61.3&38.8&51.5&57.1\\
    & &
    15   &37.5
     &39.7 &55.3&68.2&72.9&64.3&72.2&77.0&52.9&63.7&68.9\\
    & &
    50   &25.4
     &39.8&60.0&70.5&74.7&71.3&78.9&82.1&61.6&70.7&74.7\\ \hline

    \multirow{3}{*}{CurveGaussian} &
    \multirow{3}{*}{$\xmark$} &
    5   &128.7
     &46.9 &47.4&56.0&61.9&47.1&50.6&51.7&44.9&52.7&54.7\\
    & &
    15   &41.8
     &24.9 &73.7&77.5&81.2&76.9&78.4&79.9&71.5&76.6&79.0\\
    & &
    50   &27.1
     &23.8&\textbf{76.1}&81.4&83.9&81.1&84.3&85.7&\textbf{76.4}&80.7&82.8\\ \hline

    \multirow{3}{*}{SketchSplat} &
    \multirow{3}{*}{$\xmark$} &
    5   &105.2
     &155.8 &19.6&25.9&29.8&82.6&89.3&88.8&42.5&48.3&53.7\\
    & &
    15   &11.9
     &52.1 &54.7&62.0&64.6&87.7&92.8&92.2&68.6&77.0&77.5\\
    & &
    50   &16.0
     &34.0&68.4&74.1&77.3&\textbf{84.3}&87.8&89.1&74.2&79.1&81.5\\ \hline

    \multirow{2}{*}{Ours} &
    \multirow{2}{*}{$\cmark$} &
    5 &18.2
     &36.9 &43.9&71.1&77.5&51.7&82.6&89.0&47.2&76.0&82.7\\
    & &
    15 &\textbf{8.3}
     &\textbf{13.6} &73.6&\textbf{91.7}&\textbf{93.9}&76.3&\textbf{96.1}&\textbf{98.1}& 74.5&\textbf{93.8}&\textbf{96.2}\\
    \hline
\end{tabular}
}

\end{table*}

\section{Experiments}

\subsection{Dataset}

\paragraph{Wireframe-100K} We introduce {Wireframe-100K}, a large-scale multi-view dataset built upon the ABC dataset~\cite{koch2019abc} for data-driven 3D parametric curve reconstruction.
\textcolor{blue}{The dataset contains 100,000 CAD models with diverse shapes, topologies, and curve structures. For each model, we extract parametric wireframe primitives, including line segments, ellipses, and Bézier curves.}
Each object is associated with posed RGB-D views, camera parameters, and 3D parametric wireframe curves with persistent instance IDs across views.
\textcolor{blue}{Specifically, we render 50 calibrated views per model at a resolution of $512 \times 512$ and provide RGB images, foreground masks, depth maps, camera intrinsics and extrinsics, 3D parametric curve annotations, and instance-level 2D curve masks. The persistent IDs associate each 3D curve with its projections across different views, enabling learning of multi-view consistent curve grounding.}

\paragraph{Rendering Setup.}
To obtain realistic image observations, we render each CAD model using BlenderProc~\cite{Denninger2023} with randomized camera poses, HDR lighting, ground textures from CC0 Textures~\cite{cc0textures}, and object materials from Poly Haven~\cite{polyhavenPolyHaven}.
\textcolor{blue}{We additionally randomize material roughness and texture appearance to increase visual diversity. Camera poses are sampled to look toward the object center, with camera distances ranging from 1.5 to 3.5 meters and elevation angles between 25 and 65 degrees.} This rendering strategy produces realistic observations with varying illumination, material properties, and background appearance, making the training data closer to casual real-world captures.

\paragraph{Evaluation Benchmark.}
For fair comparison, we evaluate all methods on the full ABC-NEF benchmark~\cite{Ye_2023_CVPR}, which contains 115 diverse objects.  We re-render RGB images for these objects under realistic scene settings to better reflect casual real-world capture conditions. \textcolor{blue}{ABC-NEF test shapes are from ABC Chunk-0, while Wireframe-100K training shapes are drawn from Chunks 1-50, avoiding direct train/test overlap.}

\subsection{Experimental Settings}

\paragraph{Evaluation Metrics.}
We evaluate performance in terms of reconstruction accuracy, geometric compactness~\cite{Gao_2025_ICCV}, and computational efficiency. 
Following standard protocols~\cite{Li2024CVPR,Ye_2023_CVPR}, we report Accuracy (Acc.) and Completeness (Comp.) in millimeters, as well as Recall ($\mathrm{R}_{\tau}$), Precision ($\mathrm{P}_{\tau}$), and F-score ($\mathrm{F}_{\tau}$) under distance threshold $\tau$.   \textcolor{blue}{Since our method takes unposed images as input, the predicted geometry is defined up to a global similarity transformation. 
Following VGGT~\cite{wang2025vggt}, we align the predicted point cloud to the ground truth using the Umeyama algorithm~\cite{umeyama1991least}, and apply the resulting similarity transformation to the reconstructed 3D curves before evaluation. }

\paragraph{Baselines.}
We compare our method with five state-of-the-art baselines: EMAP~\cite{Li2024CVPR}, SGCR~\cite{yang2025sgcr}, EdgeGaussians~\cite{chelani2024edgegaussians}, CurveGaussian~\cite{Gao_2025_ICCV}, and SketchSplat~\cite{Ying_2025_ICCV}.

\paragraph{Implementation Details.}
We train CGGT on 4 NVIDIA A100 GPUs using the AdamW optimizer. The pretrained VGGT~\cite{wang2025vggt} backbone, camera head, and depth head are frozen, while all remaining parameters are optimized with a learning rate of $1 \times 10^{-5}$.  For each sample, we randomly select one object and sample 2 to 12 views, with a total batch size of 48 images.  We use 64 curve queries and skip objects with more than 64 curve instances during training.  Following VGGT, we apply image augmentations including random scaling, cropping, and blur. For the parametric curve optimization stage, we first back-project the grounded 2D curve pixels into 3D using the predicted depth maps and camera poses.  RANSAC-based fitting is used to initialize line, ellipse, and Bézier curve candidates.  We then jointly optimize the parameters of all curves using Adam with a learning rate of $1\times10^{-3}$. 
The optimization runs for 300 iterations and typically converges within approximately 10 seconds on a single GPU.

\begin{figure*}[t]
  \includegraphics[width=0.98\textwidth]{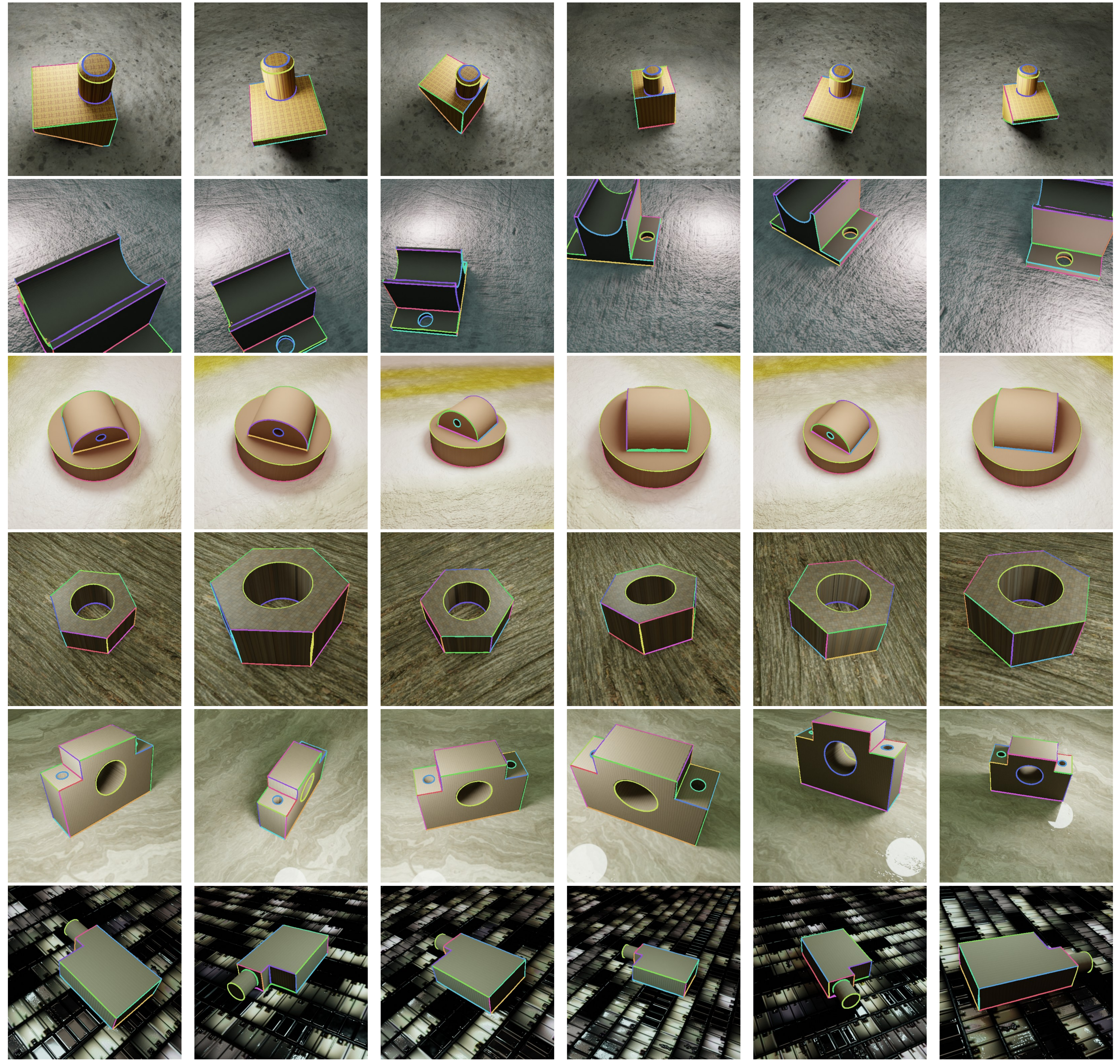}
  \vspace{-5pt}
  \caption{\textbf{Multi-view curve mask visualization.}
  We visualize the instance-level curve masks CGGT predicts from six input views. The results show that CGGT produces 3D-consistent curve masks across views, is robust to texture and silhouette edges, and generalizes well to parts with diverse shapes.}
  \label{fig:v1}
\end{figure*}

\subsection{Evaluations}
\begin{table}[tbh]
\centering
\caption{\textbf{Comparison of runtime and geometric compactness.} Our method achieves higher efficiency and produces more compact reconstructions (lowest $N_{\text{curve}}$). Besides, our approach is capable of recovering ellipses in addition to lines and B\'{e}zier curves.}
\vspace{-5pt}
\label{tab:time}
\scalebox{1.0}{
\begin{tabular}{c|c|cc}
\hline
   Method       &  Modal & Runtime $\downarrow$ & $N_{\text{curve}}$$\downarrow$  \\
\hline

EMAP  & Line, B\'{e}zier   &  3 hours &85.3 \\
EdgeGaussians    &Line, B\'{e}zier     &  4 mins &65.9 \\

SGCR   &  Line, B\'{e}zier    & 90 s   &62.7 \\
SketchSplat  & Line, B\'{e}zier & 10 mins     & 40.4 \\
CurveGaussians      & Line, B\'{e}zier                    & 5 mins    & 61.9 \\
Ours         & Line, B\'{e}zier, ellipse                &\textbf{ 10 s}     &\textbf{ 21.4} \\
\hline
\end{tabular}
}
\end{table}

\paragraph{Evaluation on Synthetic Dataset.}
Qualitative and quantitative results are shown in Fig.~\ref{fig:fig-main} and Tab.~\ref{tab:ABC}, respectively. 
Following their original settings, all baselines are evaluated with ground-truth camera poses, whereas our method directly operates on unposed RGB images. 
We evaluate the baselines with 5, 15, and 50 input views to analyze their sensitivity to view sparsity.

As shown in Fig.~\ref{fig:fig-main}, the baselines are sensitive to texture edges, illumination changes, and view-dependent silhouette contours, often producing fragmented or semantically inconsistent curves. 
In contrast, by learning instance-level curve priors from large-scale data, our method suppresses these pseudo-edges and recovers more coherent and compact curve sets under varying camera poses, lighting conditions, and surface appearances. Tab.~\ref{tab:ABC} further confirms the quantitative advantage.  Baseline reconstruction quality degrades notably as the number of views decreases, while our method maintains strong performance under sparse-view inputs and achieves clear improvements in accuracy and completeness. 
At the strictest threshold ($\tau=10$), the performance margin shrinks, mainly because our method relies on estimated poses and depths, whereas the baselines use ground-truth camera poses. Additionally, to better illustrate cross-view consistency, we visualize the 2D curves obtained through geometric warping across different views, as shown in Fig.~\ref{fig:v1}. \textcolor{blue}{Moreover, as shown in Tab.~\ref{tab:time}, our method offers a clear advantage in computational efficiency.}

\textcolor{blue}{Although ABC-NEF is a widely used benchmark containing objects with diverse shapes, we further stress-test our method on 20 held-out ABC objects, each containing more than 50 ground-truth curves. Under the same 15-view setting, our method achieves average Acc. and Comp. errors of 12.5 and 19.3, respectively. The increased errors primarily arise because denser and more complex curve configurations lead to more frequent occlusions and greater ambiguity in cross-view curve correspondences, making accurate 3D lifting and parametric fitting more challenging.}

\paragraph{Evaluation on Real-World Data.}
Reconstructing real-world mechanical parts remains challenging because casually captured images often contain sensor noise, cluttered backgrounds, and large viewpoint variations. We evaluate CGGT on real-world objects from the MV2Cyl dataset~\cite{hong2024mv2cyl}, using eight input views for each qualitative example. \textcolor{blue}{For quantitative evaluation, we follow CurveGaussian~\cite{Gao_2025_ICCV} and evaluate six real-world objects using ten input views per object. In contrast, the baseline methods use dense-view inputs following their original settings. We compare CGGT with state-of-the-art methods, including CurveGaussian and EdgeGaussians, and follow MV2Cyl to align each reconstructed edge point cloud with its ground-truth CAD model using ICP before evaluation. Since the images may contain multiple objects, we apply target-object foreground masks to exclude background clutter.}

\begin{figure}[t]
  \centering
\includegraphics[width=\linewidth]{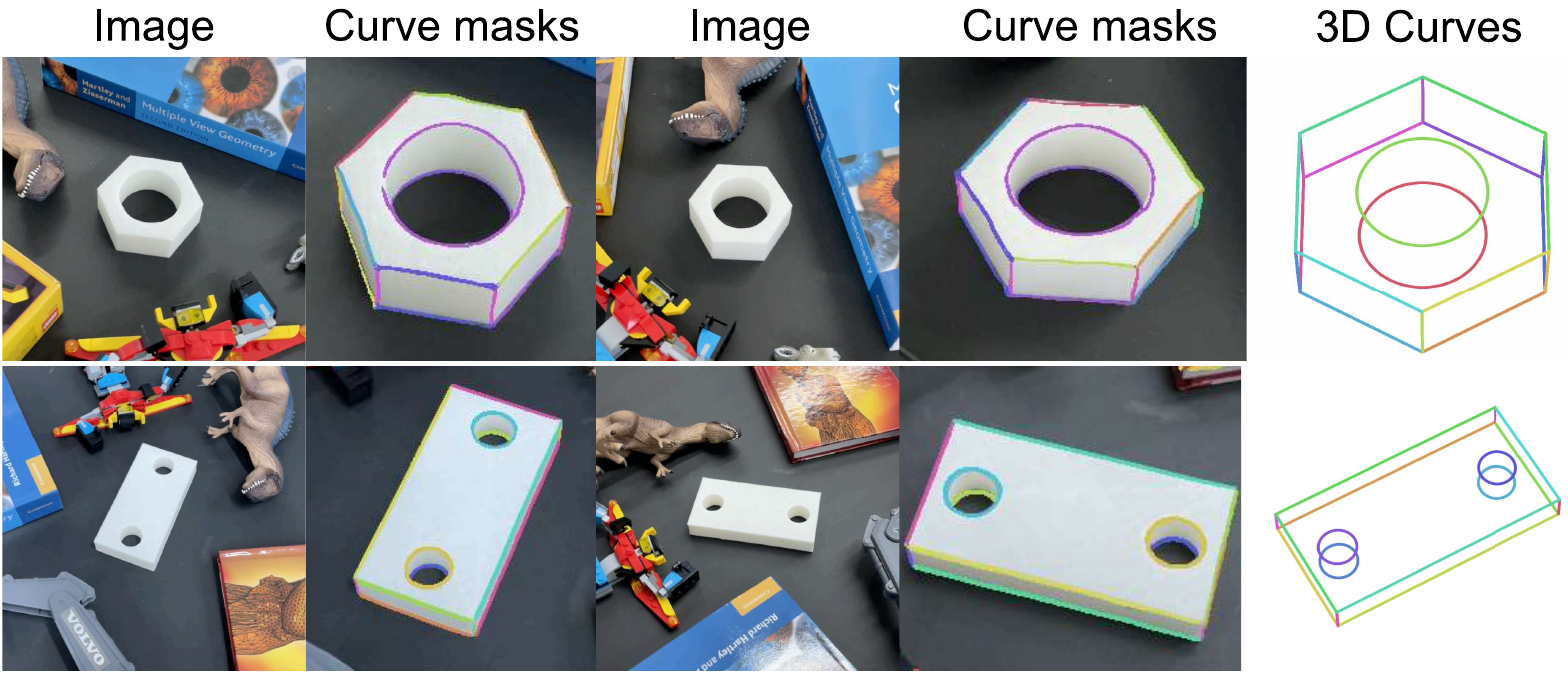}
\caption{\textbf{Qualitative results on real-world data.} Our method effectively grounds 2D curve instances and reconstructs final 3D wireframes. Object masks filter out background detections.}
  \label{fig:fig-real}
  \vspace{-2pt}
\end{figure}

As shown in Fig.~\ref{fig:fig-real} and \textcolor{blue}{Tab.~\ref{tab:real_demo}}, despite being trained solely on the synthetic \textit{Wireframe-100K} dataset, CGGT generalizes well to real images containing shadows, reflections, and distracting appearance cues. \textcolor{blue}{It also achieves higher reconstruction accuracy with substantially fewer curve primitives. Unlike optimization-based baselines, which do not explicitly establish instance-level curve correspondence across views and may produce fragmented or redundant segments, CGGT exploits learned curve-level priors and multi-view instance grounding to preserve the underlying object structure with a more compact representation.}

\begin{table}
\centering
\caption{\textcolor{blue}{\textbf{Quantitative comparisons on real objects from MV2Cyl.} Our method produces substantially more compact reconstructions using fewer 3D curve primitives.}}
\vspace{-5pt}
\scalebox{0.9}{
\begin{tabular}{c|cc|ccc|c}
\hline
   Method & Acc. $\downarrow$      & Comp. $\downarrow$ & $\text{R}_{10} \uparrow$  &  $\text{P}_{10} \uparrow$  &  $\text{F}_{10} \uparrow$  &$ N_{\text{curve}}$ $\downarrow$  \\
\hline

EdgeGaussians  &  13.8 &{7.7}& 74.4 & 46.1& 56.7 & 1952.5 \\
CurveGaussian  & {8.5}& 7.8& {74.6}&{ 71.6}& {73.1}& {70.5} \\ 

Ours & {6.3}& 5.9& {81.8}&{ 78.5}& {79.4}& {21.3} \\ 
\hline
\end{tabular}
}

\label{tab:real_demo}

\end{table}

\begin{figure}[t]
  \centering
  \includegraphics[width=1.0\linewidth]{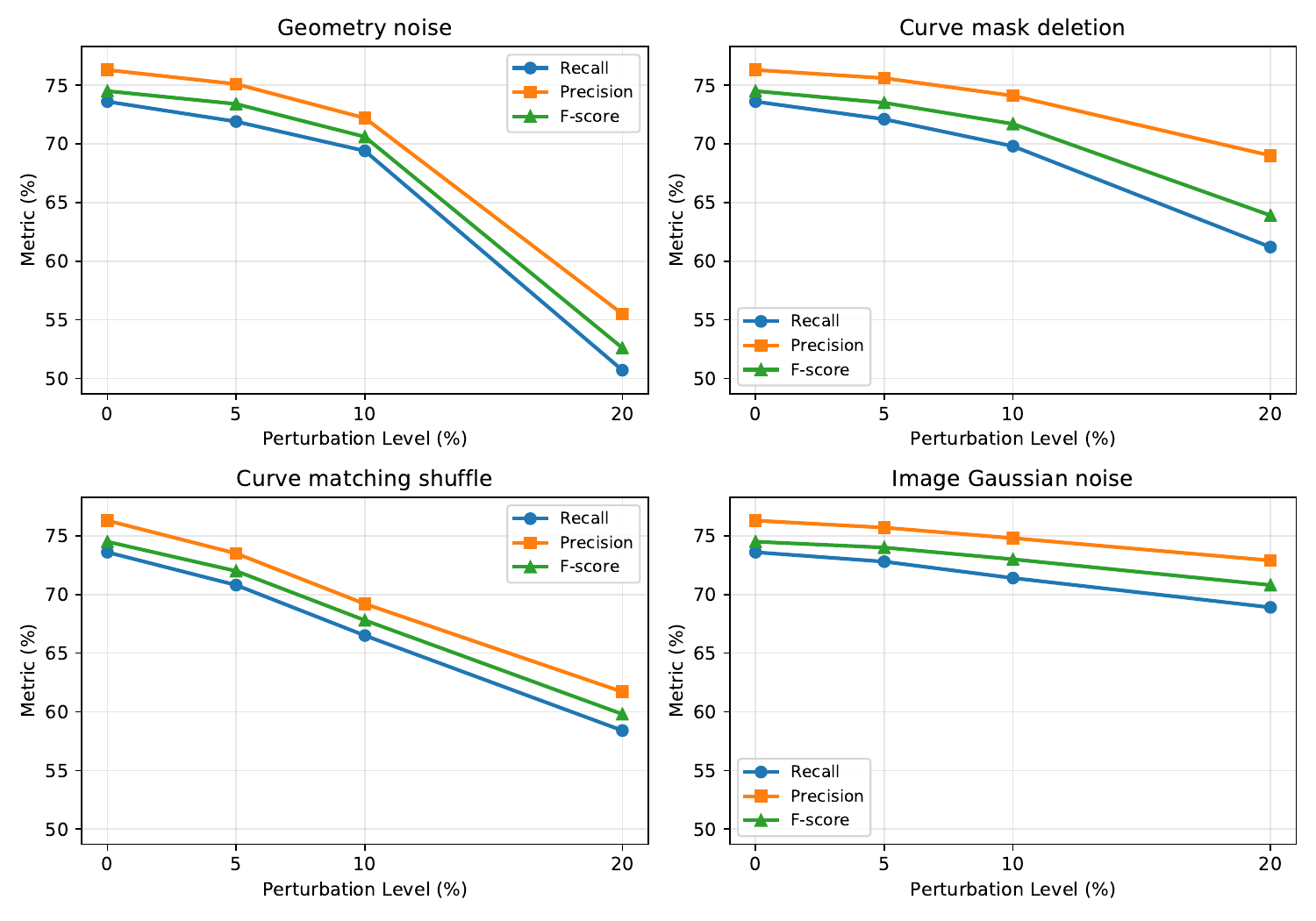}
    \vspace{-10pt}
  \caption{ \textcolor{blue}{\textbf{Robustness boundary analysis.}
  We perturb Phase I predictions and input images at 5\%, 10\%, and 20\% levels.
  Our method is relatively robust to image noise, curve mask deletion, and limited correspondence errors. Still, it is more sensitive to geometry noise because pose and depth errors directly affect 3D lifting and parametric fitting.}}
   \vspace{-3pt}
  \label{fig:robustness_boundary}
\end{figure}

\begin{table}[h]
\centering
\vspace{-1pt}
\caption{\textbf{Ablation study of key components.} We evaluate the impact of the proposed architecture components, training losses, and optimization terms on reconstruction quality.}
\vspace{-5pt}
\label{tab:abl}
\scalebox{0.9}{
\begin{tabular}{l|ccccc}
       Method & Acc.$\downarrow$ & Comp.$\downarrow$ & $\text{R}_{10} \uparrow$ & $\text{P}_{10}\uparrow$ & $\text{F}_{10} \uparrow$  \\
    \midrule

    \textbf{Ours (Full Model)} & \textbf{8.3 }& \textbf{13.6 }& \textbf{73.6} &\textbf{76.3 } &\textbf{ 74.5}   \\
    \midrule
  3D regression& 16.4& 32.2& 55.3& 60.5& 58.1  \\
Mask attention & 8.4& 19.2 & 60.4&76.1&68.6 \\
    
    w/o Depth-Aware Enhan. & 9.6 & 14.2 & 68.1 & 73.6 &   70.9 \\
    w/o Curve-Depth Cons. & 12.8 & 15.9 & 62.7 & 64.6 &   63.4 \\
    
    \midrule
    w/o Geometric Cons. &  13.6& 16.3 & 61.4 & 64.2 & 61.9   \\
    w/o Geometric Reg. & 10.6 &15.0  &70.1  &68.7  & 66.9   \\

    \bottomrule
\end{tabular}
}
\vspace{-10pt}
\end{table}

\begingroup
\color{blue}

\subsection{Robustness Analysis}

Our framework relies on first-stage CGGT predictions, whose errors may propagate to subsequent 3D lifting and parametric optimization. To better understand the robustness boundary of our method, we conduct controlled perturbation experiments on the Phase I outputs and input images. 
Starting from the original performance of the full model, i.e., $\mathrm{R}_{10}=73.6$, $\mathrm{P}_{10}=76.3$, and $\mathrm{F}_{10}=74.5$, we evaluate four types of perturbations at three levels: 5\%, 10\%, and 20\%.

Specifically, we consider: 
(1) adding geometric noise to the predicted poses and depth maps; 
(2) randomly deleting curve-mask pixels to simulate missed curve detections; 
(3) randomly shuffling curve correspondences across views to simulate instance matching errors; and 
(4) adding Gaussian noise to the input images to evaluate robustness to image degradation. 
The results are shown in Fig.~\ref{fig:robustness_boundary}.

\paragraph{Geometry noise.}
Geometry noise has a relatively strong impact on the final reconstruction quality. 
This is expected, since inaccurate poses or depth maps directly affect 3D lifting and subsequent parametric fitting. Since CGGT builds upon a 3D foundation model, it can naturally benefit from future advances in feed-forward 3D reconstruction, which would further improve the downstream 3D curve reconstruction quality.

\paragraph{Curve-mask deletion.}
Randomly deleting curve-mask pixels causes only moderate degradation, especially at low perturbation levels. 
This robustness comes from the multi-view nature of our framework: even if a curve is partially missed in some views, observations from other views can still provide sufficient evidence for 3D lifting and fitting. 
This result suggests that the parametric optimization stage can tolerate moderately incomplete masks.

\paragraph{Curve correspondence errors.}
Shuffling a small portion of curve correspondences has limited impact at 5\% and 10\%, indicating that the optimization can absorb moderate correspondence noise through multi-view consistency. 
However, the degradation becomes significant at 20\%. 
This is because incorrect cross-view curve correspondences introduce direct constraint noise into the second-stage optimization, causing unrelated 2D observations to be fitted as the same 3D curve. 
This result highlights the importance of reliable instance-level curve association.

\paragraph{Image Gaussian noise.}
Adding Gaussian noise to the input images has the mildest effect among the four perturbations. 
We attribute this robustness to CGGT's large-scale training and the image augmentations used during training, including random scaling, cropping, and blur. 
These augmentations help the model learn stable curve-level priors beyond local pixel noise.

Overall, the analysis shows that our framework is robust to moderate mask incompleteness, image degradation, and limited correspondence noise, while being more sensitive to severe geometric errors. 
This behavior is consistent with our two-stage design: the first stage focuses on image-space curve grounding, whereas the second stage relies on accurate geometry for 3D lifting and parametric refinement.

\endgroup

\subsection{Ablation Studies}

\begin{figure}[t]
  \centering
\includegraphics[width=1.0\linewidth]{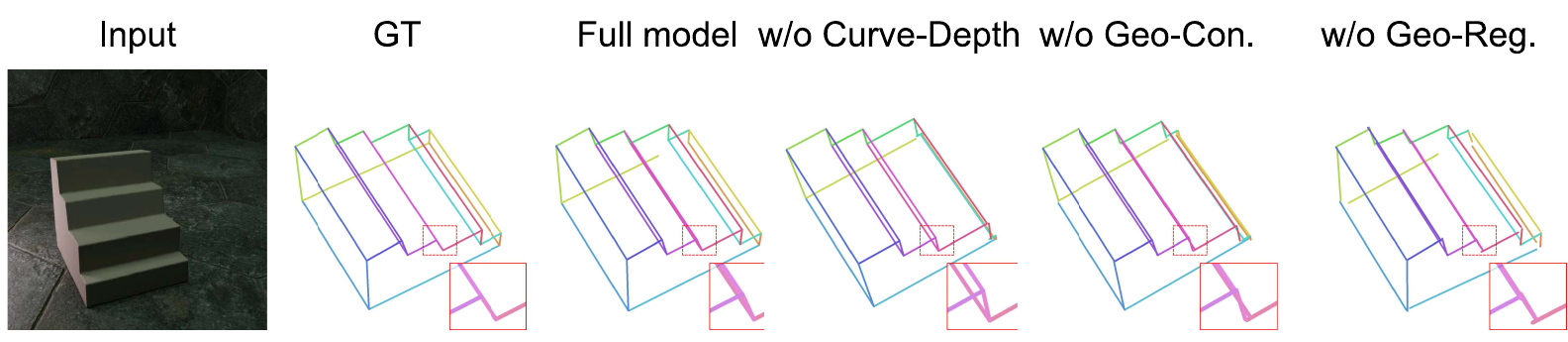}
\caption{\textbf{Visual ablation study of key components.}  As observed, omitting these critical components results in noisier reconstructions, demonstrating their necessity for achieving clean and accurate 3D parametric curves.}
  \label{fig:fig-abl}
\end{figure}

We ablate the key components in both CGGT training and test-time parametric optimization. 
Quantitative results and qualitative comparisons are shown in Tab.~\ref{tab:abl} and Fig.~\ref{fig:fig-abl}, respectively.

\paragraph{Two-stage formulation.}
Direct 3D regression leads to a large performance drop, indicating that learning 3D curve parameters directly from sparse views is highly ambiguous. 
When the input views are limited, the visible image evidence only partially constrains the full 3D curve geometry, making direct parameter regression difficult to learn. 
In contrast, curve grounding in the pixel space is better constrained by visible image observations and provides more stable supervision. 

\paragraph{Curve-aware attention.}
Replacing our curve-aware attention with standard masked attention reduces recall and F-score. 
Standard masked attention~\cite{cheng2022masked} relies on masks predicted by previous decoder layers to restrict the attention region. 
For thin curve structures, early masks are often sparse, incomplete, or noisy, causing later layers to suppress true curve pixels.

\paragraph{Depth-aware design.}
Removing either the depth-aware enhancement module or the curve-depth consistency loss degrades reconstruction quality. 
Depth features guide the decoder toward regions with sharp geometric variations, while curve-depth consistency suppresses predictions across depth-inconsistent regions. 
Together, they reduce false-positive fragments and improve both image-space grounding and subsequent 3D curve fitting.

\paragraph{Optimization objectives.}
The test-time optimization terms are essential for producing usable parametric curves.  Removing geometric consistency weakens the alignment between the fitted curves and multi-view 2D evidence. 
Removing geometric regularization yields less compact, less structured results because it no longer encourages nearby endpoints and dominant angular relationships. 

\section{Conclusion and Future Work}

We propose a two-stage framework for reconstructing editable 3D parametric curves from sparse, unposed images. CGGT first grounds curves in image space by jointly estimating geometry and multi-view consistent curve masks, followed by efficient 3D lifting and parametric refinement. Trained on the Wireframe-100K dataset, CGGT learns strong curve priors that enable accurate, efficient, and robust sparse-view reconstruction. \textcolor{blue}{Despite these advantages, CGGT relies on VGGT for initial pose and depth estimation, and geometry errors may propagate to the subsequent curve-lifting and optimization stages. Future work will explore tighter coupling between geometry estimation and curve grounding, together with stronger structural priors for resolving occlusions, symmetries, and ambiguous cross-view correspondences.}

 \section*{Acknowledgment}
 This work is supported in part by the NSFC (62325211, 62132021, 62572477, 62522219, 62372457, 62322207), the Research Fund of Jiangsu Key Laboratory of AI for Industries (E6420016G8), the Young Furong Scholar Support Program of Hunan Province, and the National Natural Science Foundation of China (62402171).

\makeatletter
\let\titleold\title
\newcommand{\thetitle}{}
\renewcommand{\title}[1]{%
  \titleold{#1}%
  \gdef\thetitle{#1}%
}
\makeatother

\def\maketitlesupplementary{%
   \newpage
   \twocolumn[
      \centering
      \Large
      \textbf{\thetitle}\\
      \vspace{0.5em}Supplementary Material of CGGT\\
      \vspace{1.0em}
   ]%
}

%
%
%
%




\bibliographystyle{ACM-Reference-Format}
\bibliography{sample-bibliography}



\end{document}